\documentclass[conference]{IEEEtran}
\IEEEoverridecommandlockouts

\usepackage{cite}
\usepackage{amsmath,amssymb,amsfonts}
\usepackage{algorithm}
\usepackage{algorithmic}
\usepackage{graphicx}
\usepackage{adjustbox}
\usepackage{bm}
\usepackage{textcomp}
\usepackage{xcolor} 
\usepackage{arydshln}
\usepackage{comment}
\usepackage{threeparttable}
\usepackage[table]{xcolor}
\usepackage{newunicodechar}
\newunicodechar{✔}{\checkmark}
\newunicodechar{✖}{\ding{55}}
\usepackage{pifont}
\usepackage{colortbl}
\usepackage{hyperref}
\hypersetup{
    hidelinks
}
\definecolor{darkgreen}{RGB}{20,100,25}

\def\BibTeX{{\rm B\kern-.05em{\sc i\kern-.025em b}\kern-.08em
    T\kern-.1667em\lower.7ex\hbox{E}\kern-.125emX}}
\begin{document}

\title{Multi-Scale Feature Attention Network for Polymer Classification Using Terahertz Spectroscopy}

\author{
\IEEEauthorblockN{Roshni Mahtani\IEEEauthorrefmark{1}, Ilán Carretero\IEEEauthorrefmark{1}, Daniel Moreno-Paris\IEEEauthorrefmark{2}, Aldo Moreno-Oyervides\IEEEauthorrefmark{2},\\ Laura Monroy\IEEEauthorrefmark{2},  Oscar Elías Bonilla-Manrique\IEEEauthorrefmark{2},  Rocío del Amor\IEEEauthorrefmark{1}\IEEEauthorrefmark{3}}
\IEEEauthorrefmark{1}\textit{Instituto Universitario de Investigación e Innovación en Tecnología Centarada en el Ser Humano,}\\ \textit{HumanTech, Universitat Politècnica de
València, Valencia, Spain}\\ 
\IEEEauthorrefmark{2}\textit{Department of Electronic Technology, Universidad Carlos III de Madrid, Leganés, Spain}\\
\IEEEauthorrefmark{3}\textit{Artikode Intelligence S.L., Valencia, Spain}\\ 
}

\maketitle

\begin{abstract}
Reliable polymer identification is essential for ensuring the quality and safety of recycled plastics, yet conventional sorting and spectroscopic techniques often struggle to deliver robust discrimination. Terahertz (THz) spectroscopy offers a promising alternative, providing high-resolution and non-destructive measurements. In this work, we leverage THz signals to classify 12 types of polymers, including pure polymers, multilayer films, commercial blends, and biopolymers. To handle the complexity of these spectral signals, we propose the Multi-Scale Feature Attention Network (MSFAN), a novel deep learning architecture tailored for THz data. The framework integrates feature gating for signal recalibration and multi-scale parallel convolutions to capture diverse frequency patterns. These features are further refined through cross-feature attention and attention pooling, enabling the model to intrinsically highlight the most informative THz regions. MSFAN consistently outperforms state-of-the-art models, reaching a classification accuracy of 85.2\%. This study demonstrates the potential of combining THz spectroscopy with deep learning techniques for effective, scalable, and interpretable polymer classification.
\end{abstract}



\section{INTRODUCTION}

Efficient plastic recycling is essential to reduce the environmental impact, promote resource circularity, and ensure product safety. However, industrial recycling is often bottlenecked by the technical difficulty of accurately classifying polymers within heterogeneous waste streams. Even minor impurities or toxic residues can compromise the mechanical integrity and safety of recycled end products, particularly in food packaging and medical devices \cite{jiang2025systematic}. 

Traditional sorting methods, such as manual inspection and chemical testing, are either limited to surface properties or destructive and environmentally hazardous \cite{mentes2023combustion}. Although spectroscopic techniques such as infrared (IR) and Raman offer non-invasive alternatives, their performance can be affected by environmental noise, weak signals, or fluorescence interference \cite{cho2003investigation}. Terahertz (THz) spectroscopy offers a promising solution due to its high penetration depth, minimal sample alteration, and the ability to measure both amplitude and phase, enabling accurate determination of absorption and refractive index \cite{tonouchi2007cutting}. 
However, the inherent complexity of the resulting spectral signals necessitates data-driven approaches to extract discriminative information for polymer characterization \cite{sarjavs2021automated}.

Deep learning (DL) has become widely adopted for automated polymer classification, achieving strong performance across a range of architectures \cite{singh2025deep}. Nonetheless, much of the existing literature relies on datasets composed exclusively of pure polymers \cite{neo2023deep,singh2023hyperspectral} or synthetically generated mixtures \cite{angulo2022machine}. Such datasets only partially reflect the material variability encountered in practice, where polymer samples may consist of blends, multilayer structures, or bio-based materials, with more complex spectral signatures.

In this work, we tackle these challenges by using THz spectroscopy measurements to classify 12 types of polymers, including pure polymers, multilayer films, commercial blends, and biopolymers. To this end, we propose the Multi-Scale Feature Attention Network (MSFAN), a novel deep learning architecture specifically designed for THz spectral analysis. The framework combines feature gating for signal recalibration, multi-scale convolutions for frequency-aware feature extraction, and cross-feature attention to refine spectral dependencies. An attention-based pooling stage further emphasizes the most informative THz regions. MSFAN achieves high accuracy in polymer discrimination while inherently extracting the most informative spectral regions. The main contributions of this study are summarized as follows:

\begin{itemize}
\item \textbf{\textit{Heterogeneous THz polymer dataset:}} We construct a dataset comprising pure polymers, multilayer films, commercial blends, and biopolymers. This design moves beyond conventional datasets dominated by pure materials, better approximating the compositional complexity encountered in recycling streams.
\item \textbf{\textit{Multi-scale attention-based architecture:}} We introduce MSFAN, a framework that integrates feature gating, multi-scale parallel convolutions, cross-feature attention, and attention-based pooling. The architecture outperforms state-of-the-art methods while inherently identifying the most informative THz frequency regions.
\end{itemize}

\begin{figure*}[htb]
    \centering
    \includegraphics[width=\textwidth]{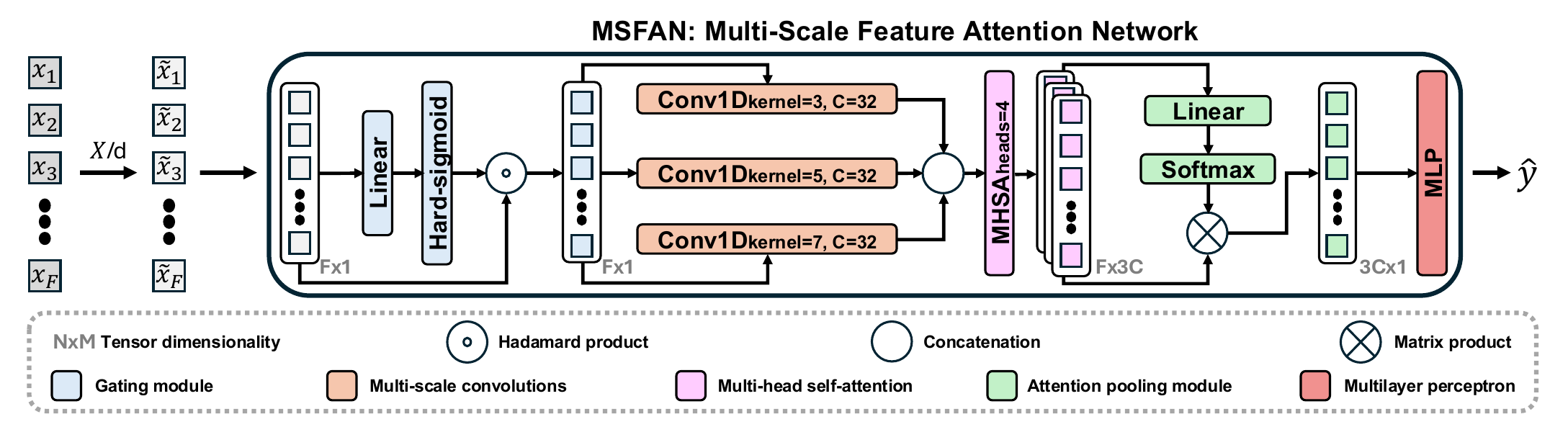}
    \caption{Overview of MSFAN architecture.
    The THz spectrum is thickness-normalized and recalibrated through a learnable gating module. Multi-scale Conv1D branches extract complementary spectral features, which are concatenated and modeled through multi-head self-attention. Softmax-based attention pooling aggregates frequency information into a compact embedding, followed by a multilayer perceptron for classification.
    }
    \label{fig:methodology}
\end{figure*}

\section{METHODOLOGY}

An overview of the proposed framework is illustrated in Fig. \ref{fig:methodology}. The following subsections provide the problem formulation and a detailed description of the architectural components.

\subsection{\textbf{Problem formulation}}  

We formulate the task as a supervised multi-class classification problem. Each spectral signature is represented as a feature vector $X = \{x_f\}_{f=1}^{F}$, consisting of $F=50$ equidistant frequency components sampled from $100$ to $590$ GHz. The dataset provides a ground-truth label $y \in \{1, \dots, Y\}$ for each sample, corresponding to one of $Y=12$ distinct polymer categories. Our goal is to learn a model $p_\theta: \mathbb{R}^F \to [0, 1]^Y$ that maps a given input $X$ to a prediction $\hat{y}$, minimizing the discrepancy with the ground-truth label $y$.

\subsection{\textbf{Physical Invariance Preprocessing}}

A fundamental challenge in transmission spectroscopy is the dependence of the spectral magnitude on the sample thickness $d$. According to the Beer-Lambert law \cite{swinehart1962beer}, the detected signal intensity is inherently coupled to the optical path length, acting as a scaling factor extrinsic to the material's chemical identity. To mitigate this variability without requiring absolute transmission references and enforce geometric invariance, we implement a normalization strategy motivated by physical principles. Adopting a first-order approximation of the attenuation process, we define a thickness-invariant feature space $\tilde{X}$ by normalizing the raw spectral amplitudes relative to the sample thickness: $\tilde{x}_f = x_f / d$. This simple yet effective transformation eliminates the geometric bias introduced by thickness variations, ensuring that the predictive model $p_\theta$ relies solely on the intrinsic spectral signature of the polymer.

\subsection{\textbf{MSFAN: Multi-Scale Feature Attention Network}}

\subsubsection{\textbf{Spectral Recalibration and Multi-Scale Extraction}}

To disentangle subtle spectroscopic patterns from the thickness-invariant input $\tilde{X} \in \mathbb{R}^{F}$, we initiate the pipeline with a learnable spectral recalibration module. We first project the normalized spectrum through a dense gating layer parameterized by weights $W_g \in \mathbb{R}^{F \times F}$ and bias $b_g \in \mathbb{R}^{F}$. Activated by a hard-sigmoid function $\sigma_{HS}$, this layer generates an attention mask $\boldsymbol{\alpha} = \sigma_{HS}(W_g \tilde{X} + b_g)$, which is applied element-wise ($X_{g} = \tilde{X} \odot \boldsymbol{\alpha}$) to dynamically reweight the spectral bands, suppressing background noise while emphasizing discriminative THz features.

The recalibrated signal is subsequently processed by an inception-inspired multi-scale block designed to capture morphological variations at diverse resolutions. We employ three parallel 1D convolutional branches with varying kernel sizes ($\text{kernel} \in \{3, 5, 7\}$), each composed of $C=32$ filters. This parallel topology enables the simultaneous encoding of high-frequency local peaks and broad spectral trends. The branch outputs are passed through ReLU activations and concatenated along the channel dimension, yielding a rich feature map $S \in \mathbb{R}^{3C \times F}$. Finally, batch normalization is applied to stabilize the distribution of these features.

\subsubsection{\textbf{Global Spectral Contextualization and Weighted Aggregation}}  

While the convolutional stage extracts local morphological features, it lacks a global receptive field. To capture long-range dependencies, we treat the spectral feature map as a sequence $S^{\mathrm{T}}
 \in \mathbb{R}^{F \times 3C}$ and apply a multi-head self-attention (MHSA) mechanism. We first project the input into queries ($Q$), keys ($K$), and values ($V$) via learnable matrices. The output context $Z  \in \mathbb{R}^{F \times 3C}$ is generated by computing the alignment scores between spectral bands:
$Z = \text{softmax}\left(\frac{QK^\top}{\sqrt{d_k}}\right)V$, where ${d_k}$ represents the dimension of $K$ and $Q$.

Subsequently, to condense this sequence into a fixed-length descriptor, we implement a content-aware attention pooling mechanism. We compute a scalar importance score for each of the $F$ spectral frequencies via a linear projection and softmax normalization, generating a weight vector $a \in \mathbb{R}^{F}$. The final embedding $H \in \mathbb{R}^{3C}$ is derived via a weighted aggregation $H = \sum_{j=1}^{F} a_j z_{j}$, where $z_j$ denotes the $j$-th row vector of $Z$. This formulation ensures that the feature representation is explicitly driven by the most discriminative signals prior to classification.

\subsubsection{\textbf{Classification Head and Sparse Optimization}} 

The aggregated spectral descriptor $H$ is mapped to the final decision space through a multilayer perceptron (MLP). This projection head comprises two fully connected hidden layers equipped with ReLU activations to capture non-linear decision boundaries. To enhance generalization, we apply dropout regularization prior to the penultimate layer. The final linear transformation yields the class logits $\hat{y} \in \mathbb{R}^{Y}$, representing the unnormalized log-probabilities for the polymer categories.

To optimize the network parameters $\theta$, we formulate the objective function as a weighted combination of cross-entropy loss $\mathcal{L}_{{CE}}$ and an $L_1$ sparsity penalty. The primary supervision signal is provided by $\mathcal{L}_{{CE}}$ with label smoothing ($\epsilon=0.1$), which penalizes misclassifications while preventing model overconfidence on noisy spectral data. Complementarily, we apply an $L_1$ regularization term to the gating activation vector $\alpha$. This sparsity constraint minimizes the cumulative magnitude of the attention weights, forcing the network to actively suppress irrelevant spectral bands and focus computational resources solely on discriminative frequencies. Therefore, MSFAN loss is defined as:

\begin{equation}\mathcal{L}_{MSFAN} = \mathcal{L}_{{CE}}(\hat{y}, y_{smooth}) + \lambda \frac{1}{F} \sum_{k=1}^{F} |\alpha_k|.\end{equation}

\vspace{1mm}
\section{EXPERIMENTAL SETTING}

\subsection{\textbf{Dataset}}

The dataset comprises THz spectroscopy measurements collected over five consecutive days from 12 polymer types, shown in Table \ref{tab:ddbb}. Materials include pure polymers (D, I, J, L, O), multilayer structures (A, B, F, H), commercial blends (C), and biopolymers (E, G). For each material, signals were acquired in both low-gain (LG) and high-gain (HG) configurations to ensure a wide dynamic range across the terahertz spectrum. 

\begin{table}[ht]
\centering
\caption{Polymer samples and corresponding thicknesses}
\label{tab:ddbb}
\renewcommand{\arraystretch}{1.2}
\resizebox{\columnwidth}{!}{
\begin{tabular}{lll}
\hline
\textbf{ID} & \textbf{Material} & \textbf{Thickness (mm)} \\
\hline
A & PE/tie/EVOH/tie/PE/Adhesive/PE/tie/EVOH/tie/PE & 0.20\\
B & PE/tie/EVOH/tie/PE (Admer AT1707E) & 0.57 \\
C & ABS+PC & 2.05 \\
D & ABS & 3.00 \\
E & Ecovio/PVOH/Ecovio & 0.10 \\
F & PP/tie/EVOH/tie/PP (tupper) & 0.29 \\
G & PHB/PVOH/Ecovio & 0.10 \\
H & PP/tie/EVOH/tie/PP & 0.07 \\
I & PS & 0.36 \\
J & LDPE & 0.07 \\
L & PVC & 1.85 \\
O & PET & 0.12 \\
\hline
\end{tabular}
}

\vspace{1mm}
{\scriptsize
PE: polyethylene, tie: adhesive layer, EVOH: ethylene vinyl alcohol, ABS: acrylonitrile butadiene styrene, PC: polycarbonate, Ecovio: biodegradable blend, PVOH: polyvinyl alcohol, PP: polypropylene, PHB: polyhydroxybutyrate, PS: polystyrene, LDPE: low-density polyethylene, PVC: polyvinyl chloride, PET: polyethylene terephthalate.
}
\end{table}

The measurement protocol operated in the terahertz region (100–590 GHz) with a 10 GHz step, yielding $F=50$ discrete frequency points. At each frequency step, the system recorded the polymer response for 12 seconds, capturing multiple temporal snapshots.

To construct the final spectral feature vectors, we applied a temporal alignment strategy across frequency steps. Specifically, the $i$-th spectral instance is generated by concatenating the $i$-th temporal acquisition snapshot from every frequency channel. To ensure a strictly balanced benchmark, we identified the minimum number of valid snapshots common to all experiments, retaining $N=209$ independent spectral instances per polymer, per day, and per gain configuration. Consequently, the resulting dataset inherently incorporates physical variability, including sample thicknesses ($d \in [0.07, 3.00]$ mm) and environmental fluctuations, namely temperature ($T \in [19.7, 24.3]^\circ$C) and relative humidity ($RH \in [25.4, 42.6]\%$). To support reproducibility and facilitate future research, the processed dataset is openly accessible at \href{https://github.com/danimp94/PLASTICS-THz/tree/main/data/experiment_5_plastics}{https://github.com/danimp94/PLASTICS-THz}.

\subsection{\textbf{Signal integration strategies}}

To exploit the complementary dynamic ranges of LG and HG signals, we investigate distinct processing configurations. To ensure scale consistency, we apply independent min-max normalization to each signal. We benchmark single-source performance (LG, HG) against input-level aggregation (early fusion), including element-wise averaging (MEAN), maximization (MAX), and channel concatenation (CONCAT). Additionally, we introduce late fusion aggregation (DUAL), in which independent DL-based backbones process each gain stream separately, concatenating their latent representations prior to final classification.

\subsection{\textbf{Experimental configuration}}

We validate the proposed framework using a leave-one-day-out cross-validation scheme to rigorously assess generalization against temporal variability, effectively constituting a stratified 5-fold protocol. Model performance is quantified by accuracy (ACC), specificity (SPE), precision (PPV), and F1-Score (F1), noting that ACC is mathematically equivalent to macro-averaged recall for our balanced dataset. Parameters are optimized end-to-end over 100 epochs using the Adam solver (initial learning rate $1 \times 10^{-4}$, batch size 128), minimizing the joint objective function $\mathcal{L}_{\text{MSFAN}}$, with convergence stabilized via \textit{ReduceLROnPlateau} scheduler. Source code is publicly available at \href{https://github.com/roshni-mahtani/MSFAN}{https://github.com/roshni-mahtani/MSFAN}.

\subsection{\textbf{State-of-the-art architectures}}

To establish a comprehensive benchmark, diverse deep learning architectures previously employed in state-of-the-art spectroscopic polymer classification were evaluated. A standard artificial neural network (ANN) \cite{singh2023hyperspectral}, comprising a feedforward MLP with batch normalization, serves as the dense baseline, while a convolutional neural network (CNN) \cite{ng2019convolutional} composed of a four-stage hierarchy of convolutions and max-pooling operations represents the convolutional baseline. The PSDN\_Inception\cite{neo2023deep} was also included to evaluate multi-scale feature extraction capabilities. Furthermore, advanced architectures integrating a trainable preprocessing module (PM) designed to stabilize inputs via learnable pooling and normalization were assessed. Specifically, ANN\_PM\cite{singh2025deep} incorporates this module prior to the dense layers to enhance feature robustness. The Improved\_CNN\_PM\cite{singh2025deep} refines the Inception architecture by introducing dilated convolutions to expand the receptive field. Finally, Transformer\_PM\cite{singh2025deep} exploits self-attention mechanisms to capture global spectral dependencies.

\section{RESULTS}

\subsection{\textbf{Analysis of thickness invariance and signal aggregation}}

\begin{figure}[htb]
    \centering
    \includegraphics[width=1\linewidth]{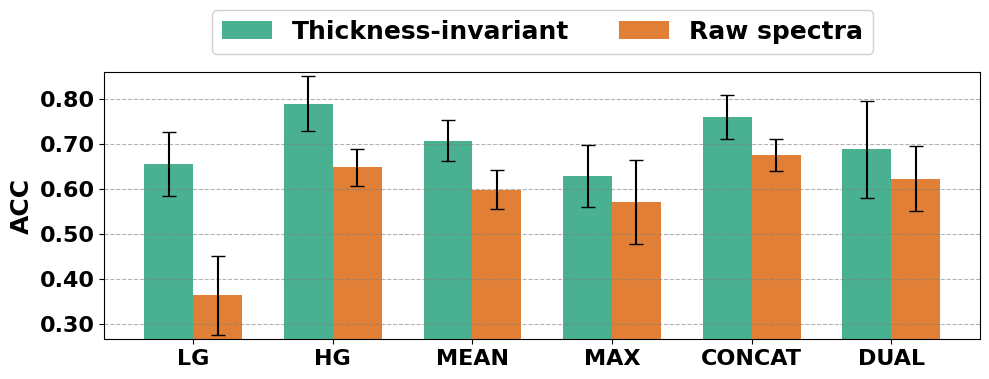}
    \caption{Mean classification accuracy obtained with different signal aggregation strategies across the evaluated DL models. Results are shown for thickness-invariant and raw spectral inputs. Error bars denote standard deviation.}
    \label{fig:models_agg}
\end{figure}

We initiate our empirical analysis by evaluating the optimal input configuration and reporting the mean accuracy across all evaluated deep learning models, including state-of-the-art architectures and MSFAN. Additionally, we quantify the standard deviation to assess the consistency of the signal integration strategies independent of the downstream model. Fig. \ref{fig:models_agg} illustrates the comparative results. Applying thickness-invariant normalization yields substantial performance enhancement, outperforming raw spectral inputs by 16.05\% on average across all integration strategies. Regarding signal fusion, the isolated HG configuration achieves the highest performance, reaching an average accuracy of 79.23\%. The consistent performance degradation observed when incorporating LG spectra indicates that the low-gain stream provides negligible complementary discriminative information and potentially introduces noise. Consequently, we adopt the thickness-invariant HG signal for subsequent experiments.

\subsection{\textbf{Comparison to the literature}}

We benchmark the proposed MSFAN against state-of-the-art deep learning architectures for polymer classification, with quantitative results detailed in Table \ref{tab:model_comparison}. Our framework consistently outperforms competing approaches, obtaining 85.2\% accuracy and 82.4\% F1-score. MSFAN surpasses the strongest competitor, the Improved\_CNN\_PM \cite{singh2025deep}, by a margin of 1.4\% in accuracy and 1.3\% in F1-score. Significantly, this performance enhancement is attributed to the architecture's superior ability to capture subtle spectral distinctions between polymers. This capability is exemplified by the discrimination between polymer H (PP/tie/EVOH/tie/PP) and J (LDPE). While the Improved\_CNN\_PM achieves a pairwise accuracy of 61.1\%, MSFAN improves it to 75.1\%.

\begin{table}[htb]
\centering
\caption{Performance comparison of MSFAN and benchmark models}
\label{tab:model_comparison}
\renewcommand{\arraystretch}{1.2}
\resizebox{\columnwidth}{!}{
\begin{tabular}{lcccc}
\hline
\textbf{Model} & \textbf{ACC} & \textbf{SPE} & \textbf{PPV} & \textbf{F1} \\
\hline
CNN \cite{ng2019convolutional} {\tiny\textcolor{gray}{Geoderma '19}} & 0.821 $\pm$ 0.090 & 0.984 $\pm$ 0.008 & 0.803 $\pm$ 0.103 & 0.789 $\pm$ 0.093 \\

PSDN\_Inception \cite{neo2023deep} {\tiny\textcolor{gray}{RCR '23}} & 0.704 $\pm$ 0.116 & 0.973 $\pm$ 0.011 & 0.678 $\pm$ 0.144 & 0.654 $\pm$ 0.135 \\

ANN \cite{singh2023hyperspectral} {\tiny\textcolor{gray}{PSEP '23}} & 0.703 $\pm$ 0.081 & 0.973 $\pm$ 0.007 & 0.679 $\pm$ 0.094 & 0.648 $\pm$ 0.082 \\

ANN\_PM \cite{singh2025deep} {\tiny\textcolor{gray}{JCP '25}} & 0.815 $\pm$ 0.045 & 0.983 $\pm$ 0.004 & 0.812 $\pm$ 0.053 & 0.787 $\pm$ 0.056 \\

Transformer\_PM \cite{singh2025deep} {\tiny\textcolor{gray}{JCP '25}} & 0.794 $\pm$ 0.063 & 0.981 $\pm$ 0.006 & 0.787 $\pm$ 0.077 & 0.757 $\pm$ 0.085 \\

Improved\_CNN\_PM \cite{singh2025deep} {\tiny\textcolor{gray}{JCP '25}} & 0.838 $\pm$ 0.074 & 0.985 $\pm$ 0.007 & 0.825 $\pm$ 0.090 & 0.811 $\pm$ 0.092 \\

\rowcolor{gray!15}
MSFAN \textit{(Ours)} & \textbf{0.852 $\pm$ 0.075} & \textbf{0.987 $\pm$ 0.007} & \textbf{0.827 $\pm$ 0.095} & \textbf{0.824 $\pm$ 0.093} \\
\hline
\end{tabular}
}

\vspace{1mm}
{\scriptsize
Values are reported as mean $\pm$ standard deviation across cross-validation folds. Boldface denotes the best result per metric; the proposed method is shaded in gray.
}
\end{table}

\subsection{\textbf{Sensitivity Analysis of the Regularization Strength}}

\begin{figure}[htb]
    \centering
    \includegraphics[width=1\linewidth]{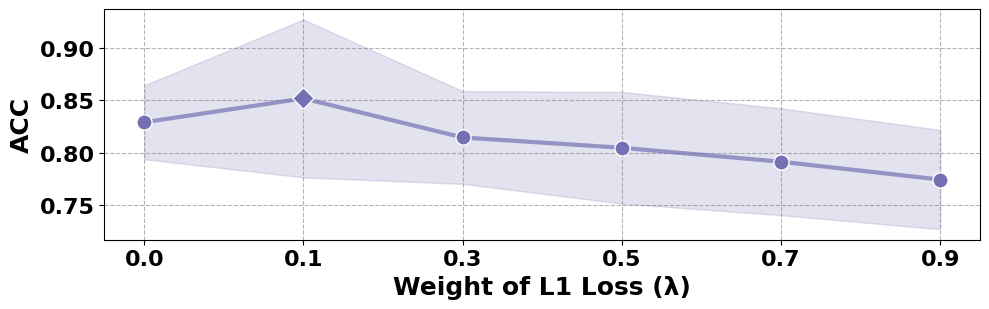}
    \caption{Ablation study on the effect of the L1 loss weight ($\lambda$) on model performance. The solid line represents the mean accuracy across folds, while the shaded area shows the standard deviation.}
    \label{fig:l1_ablation}
\end{figure}

Calibrating the regularization strength $\lambda$ is critical for balancing the classification objective with spectral sparsity. Fig.\ref{fig:l1_ablation} illustrates the impact of varying $\lambda$, identifying $\lambda=0.1$ as the optimal configuration with an accuracy of 85.2\%. This performance surpasses the unregularized baseline ($\lambda=0.0$, 82.9\%), indicating that moderate sparsity aids in isolating discriminative frequencies while suppressing noise. Conversely, increasing $\lambda$ beyond this point causes a steady performance degradation, dropping to 77.4\% at $\lambda=0.9$. This trend suggests that, while distinct spectral selection is beneficial, an overly aggressive penalty suppresses informative features essential for accurate classification.

\subsection{\textbf{Ablation Study on Architectural Components}}

To analyze individual contributions, we decouple our proposed framework into its core components: spectral gating (G), multi-scale convolutions (C), self-attention (A), attention pooling (P), and the MLP classifier (MLP). We define the baseline as the combination of P+MLP, which is applied directly to the thickness-invariant HG spectrum. Table \ref{tab:ablation_spectralcnn} quantifies the impact of progressively integrating these mechanisms. Empirical results demonstrate that combining multiple modules generally yields higher metric gains compared to isolated additions. Significantly, the full MSFAN architecture achieves the highest performance, delivering a 13.9\% absolute improvement in accuracy and a 17.3\% increase in F1-score over the baseline. This outcome validates the need for the proposed mechanisms to resolve complex polymer classification.

\begin{table}[htb]
\centering
\caption{Ablation study of MSFAN modules}
\label{tab:ablation_spectralcnn}
\renewcommand{\arraystretch}{1.2}
\resizebox{\columnwidth}{!}{
\begin{tabular}{lccccc cccc}
\hline
\textbf{G} & \textbf{C} & \textbf{A} & \textbf{P+MLP}
& \textbf{ACC} & \textbf{SPE} & \textbf{PPV} & \textbf{F1} \\
\hline

✖ & ✖ & ✖ & ✔
& 0.713 & 0.974 & 0.633 & 0.651 \\

✔ & ✖ & ✖ & ✔
& 0.725 {\tiny\color{darkgreen}{$\uparrow$ 0.012}}
& 0.975 {\tiny\color{darkgreen}{$\uparrow$ 0.001}}
& 0.664 {\tiny\color{darkgreen}{$\uparrow$ 0.031}}
& 0.668 {\tiny\color{darkgreen}{$\uparrow$ 0.017}} \\

✖ & ✔ & ✖ & ✔
& 0.837 {\tiny\color{darkgreen}{$\uparrow$ 0.124}}
& 0.985 {\tiny\color{darkgreen}{$\uparrow$ 0.011}}
& 0.817 {\tiny\color{darkgreen}{$\uparrow$ 0.184}}
& 0.811 {\tiny\color{darkgreen}{$\uparrow$ 0.160}} \\

✖ & ✖ & ✔ & ✔
& 0.755 {\tiny\color{darkgreen}{$\uparrow$ 0.042}}
& 0.978 {\tiny\color{darkgreen}{$\uparrow$ 0.004}}
& 0.746 {\tiny\color{darkgreen}{$\uparrow$ 0.113}}
& 0.707 {\tiny\color{darkgreen}{$\uparrow$ 0.056}} \\

✖ & ✔ & ✔ & ✔
& 0.776 {\tiny\color{darkgreen}{$\uparrow$ 0.063}}
& 0.980 {\tiny\color{darkgreen}{$\uparrow$ 0.006}}
& 0.751 {\tiny\color{darkgreen}{$\uparrow$ 0.118}}
& 0.737 {\tiny\color{darkgreen}{$\uparrow$ 0.086}} \\

✔ & ✖ & ✔ & ✔
& 0.807 {\tiny\color{darkgreen}{$\uparrow$ 0.094}}
& 0.983 {\tiny\color{darkgreen}{$\uparrow$ 0.009}}
& 0.774 {\tiny\color{darkgreen}{$\uparrow$ 0.141}}
& 0.775 {\tiny\color{darkgreen}{$\uparrow$ 0.124}} \\

✔ & ✔ & ✖ & ✔
& 0.826 {\tiny\color{darkgreen}{$\uparrow$ 0.113}}
& 0.984 {\tiny\color{darkgreen}{$\uparrow$ 0.010}}
& 0.809 {\tiny\color{darkgreen}{$\uparrow$ 0.176}}
& 0.796 {\tiny\color{darkgreen}{$\uparrow$ 0.145}} \\

\rowcolor{gray!15}
✔ & ✔ & ✔ & ✔
& \textbf{0.852} {\tiny\color{darkgreen}{$\uparrow$ 0.139}}
& \textbf{0.987} {\tiny\color{darkgreen}{$\uparrow$ 0.013}}
& \textbf{0.827} {\tiny\color{darkgreen}{$\uparrow$ 0.194}}
& \textbf{0.824} {\tiny\color{darkgreen}{$\uparrow$ 0.173}} \\
\hline
\end{tabular}
}

\vspace{1mm}
{\scriptsize
G = Gating (recalibration module); C = Convolutions (Multi-scale block);

A = Attention (MHSA); P+MLP = Pooling + MLP Classifier.

✔ Active module, ✖ Inactive module. Mean performance across folds is reported.

Values in green ($\uparrow$) indicate the improvement over the P+MLP baseline.
}
\end{table}

\subsection{\textbf{Spectral Saliency and Feature Attribution}}

\begin{figure}[htb]
    \centering
    \includegraphics[width=0.95\linewidth]{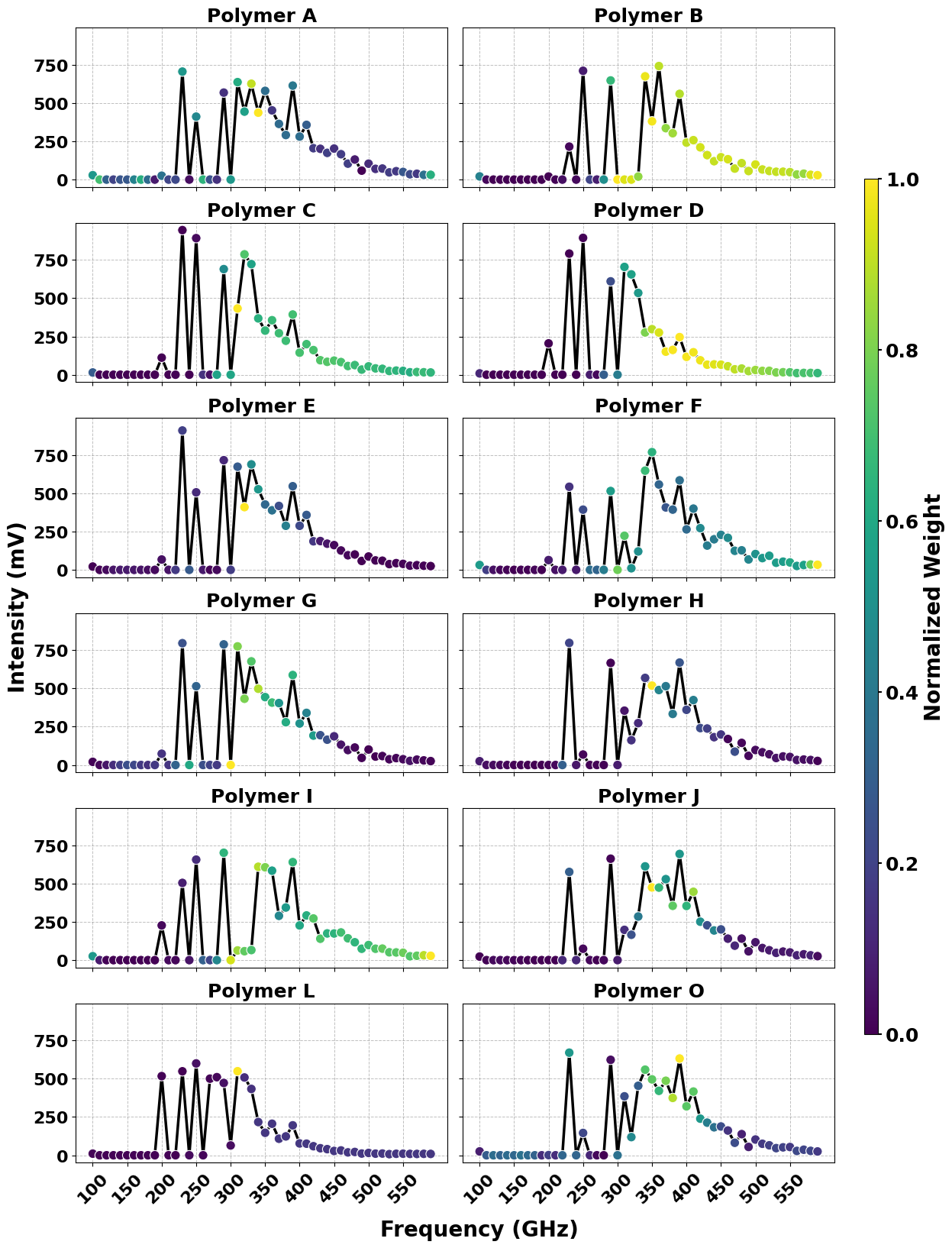}
    \caption{Spectral Attention Analysis. Mean spectral intensity profiles per polymer overlaid with normalized attention weights.}
    \label{fig:freqs}
\end{figure}

Leveraging the interpretability of the MSFAN architecture, we extract the spectral attention vector $a \in \mathbb{R}^F$ to quantify the contribution of specific frequency bands. Fig. \ref{fig:freqs} visualizes the mean spectral intensity for each polymer, overlaid with color-coded importance maps derived from the min-max-normalized average attention weights. The resulting attention profiles reveal that the model adopts distinct recognition strategies depending on the polymer structure. For Polymers A, G, J, and O, the network prioritizes the 300-400 GHz range, aligning with regions characterized by pronounced spectral fluctuations. In contrast, Polymers B, C, D, F, and I exhibit sustained attention in the high-frequency spectrum ($>400$ GHz). Finally, Polymers E, H, and L demonstrate highly localized attention to specific narrow-band frequencies. This behavior indicates that MSFAN learns to identify discriminative frequency patterns specific to each polymer category, ensuring robust performance in THz-spectroscopy-based classification.

\section{CONCLUSION}

Efficient plastic sorting is essential to ensure the quality and safety of recycled materials. To address this challenge, we used THz spectroscopy to classify 12 polymer types, including pure resins, multilayer films, commercial blends, and biopolymers. We introduced MSFAN, a deep learning framework that integrates feature gating, multi-scale spectral filtering, cross-feature attention, and attention pooling to handle the complexity of THz signals. Our results demonstrate that MSFAN consistently outperforms state-of-the-art methods, reaching a classification accuracy of 85.2\% while providing interpretable insights into the most discriminative frequency bands for each polymer class.

Despite its effectiveness, MSFAN relies on the full THz spectrum and operates under a closed-set assumption on uncontaminated samples. Real-world recycling streams, however, frequently introduce out-of-distribution materials and severe surface contamination. Future work will explore hard sparsity-driven selection to strictly isolate the most discriminative frequency bands. Furthermore, we will extend the framework with open-set recognition capabilities and evaluate its robustness against complex industrial contaminants.

\section*{ACKNOWLEDGMENT}

This research was funded by the Generalitat Valenciana (GVA) under project CIGE/2024/147 (CLAIRE).

\bibliographystyle{ieeetr}
\bibliography{bibliography}

@article{jiang2025systematic,
  title={A systematic review of plastic recycling: technology, environmental impact and economic evaluation},
  author={Jiang, Xiaoli and Bateer, Buhe},
  journal={Waste Management \& Research},
  pages={0734242X241310658},
  year={2025},
  publisher={SAGE Publications Sage UK: London, England}
}

@article{mentes2023combustion,
  title={Combustion behaviour of plastic waste--A case study of PP, HDPE, PET, and mixed PES-EL},
  author={Mentes, D{\'o}ra and Nagy, G{\'a}bor and Szab{\'o}, Tam{\'a}s J and Horny{\'a}k-Mester, Enik{\H{o}} and Fiser, B{\'e}la and Viskolcz, B{\'e}la and P{\'o}liska, Csaba},
  journal={Journal of Cleaner Production},
  volume={402},
  pages={136850},
  year={2023},
  publisher={Elsevier}
}

@article{cho2003investigation,
  title={Investigation of chemometric calibration performance based on different chemical matrix and signal-to-noise ratio},
  author={Cho, Soohwa and Chung, Hoeil},
  journal={Analytical sciences},
  volume={19},
  number={9},
  pages={1327--1329},
  year={2003},
  publisher={Springer}
}

@article{neo2023deep,
  title={Deep learning for chemometric analysis of plastic spectral data from infrared and Raman databases},
  author={Neo, Edward Ren Kai and Low, Jonathan Sze Choong and Goodship, Vannessa and Debattista, Kurt},
  journal={Resources, Conservation and Recycling},
  volume={188},
  pages={106718},
  year={2023},
  publisher={Elsevier}
}

@article{singh2025deep,
  title={Deep learning-based plastic classification using spectroscopic data},
  author={Singh, Aru Ranjan and Neo, Edward Ren Kai and Lai, Car Man and Hazra, Sumit and Coles, Stuart and Peijs, Ton and Debattista, Kurt},
  journal={Journal of Cleaner Production},
  volume={530},
  pages={146793},
  year={2025},
  publisher={Elsevier}
}

@article{singh2023hyperspectral,
  title={Hyperspectral imaging-based classification of post-consumer thermoplastics for plastics recycling using artificial neural network},
  author={Singh, Mukesh Kumar and Hait, Subrata and Thakur, Atul},
  journal={Process Safety and Environmental Protection},
  volume={179},
  pages={593--602},
  year={2023},
  publisher={Elsevier}
}

@article{ng2019convolutional,
  title={Convolutional neural network for simultaneous prediction of several soil properties using visible/near-infrared, mid-infrared, and their combined spectra},
  author={Ng, Wartini and Minasny, Budiman and Montazerolghaem, Maryam and Padarian, Jose and Ferguson, Richard and Bailey, Scarlett and McBratney, Alex B},
  journal={Geoderma},
  volume={352},
  pages={251--267},
  year={2019},
  publisher={Elsevier}
}

@article{angulo2022machine,
  title={Machine learning enhanced spectroscopic analysis: towards autonomous chemical mixture characterization for rapid process optimization},
  author={Angulo, Andrea and Yang, Lankun and Aydil, Eray S and Modestino, Miguel A},
  journal={Digital Discovery},
  volume={1},
  number={1},
  pages={35--44},
  year={2022},
  publisher={Royal Society of Chemistry}
}

@article{tonouchi2007cutting,
  title={Cutting-edge terahertz technology},
  author={Tonouchi, Masayoshi},
  journal={Nature photonics},
  volume={1},
  number={2},
  pages={97--105},
  year={2007},
  publisher={Nature Publishing Group UK London}
}

@article{swinehart1962beer,
  title={The beer-lambert law},
  author={Swinehart, Donald F},
  journal={Journal of chemical education},
  volume={39},
  number={7},
  pages={333},
  year={1962},
  publisher={ACS Publications}
}

@article{sarjavs2021automated,
  title={Automated inorganic pigment classification in plastic material using terahertz spectroscopy},
  author={Sarja{\v{s}}, Andrej and Pongrac, Bla{\v{z}} and Gleich, Du{\v{s}}an},
  journal={Sensors},
  volume={21},
  number={14},
  pages={4709},
  year={2021},
  publisher={MDPI}
}

\end{document}